\documentclass{article}

\usepackage{graphicx} % more modern
\usepackage{subfigure}
\usepackage{amssymb}
\usepackage{amsmath}
\usepackage{times}
\usepackage{helvet}
\usepackage{courier}
\usepackage{algorithm}
\usepackage{algorithmic}
\usepackage{array}
\usepackage{multirow}
\usepackage{threeparttable}
\usepackage{hyperref}

\usepackage{icml2014}
\icmltitlerunning{Copula Mixed-Membership Stochastic Blockmodel with Subgroup Correlation}

\begin{document}

\twocolumn[
\icmltitle{Copula Mixed-Membership Stochastic Blockmodel with Subgroup Correlation}

% It is OKAY to include author information, even for blind
% submissions: the style file will automatically remove it for you
% unless you've provided the [accepted] option to the icml2013
% package.
\icmlauthor{Your Name}{email@yourdomain.edu}
\icmladdress{Your Fantastic Institute,
            314159 Pi St., Palo Alto, CA 94306 USA}
\icmlauthor{Your CoAuthor's Name}{email@coauthordomain.edu}
\icmladdress{Their Fantastic Institute,
            27182 Exp St., Toronto, ON M6H 2T1 CANADA}

% You may provide any keywords that you
% find helpful for describing your paper; these are used to populate
% the "keywords" metadata in the PDF but will not be shown in the document
\icmlkeywords{boring formatting information, machine learning, ICML}

\vskip 0.3in
]

\begin{abstract}

The Mixed-Membership Stochastic Blockmodel (MMSB) is a popular framework for modeling social network relationships which fully exploits each individual node participation (or membership) in a social structure. Despite its powerful representations, this model makes an assumption that the distributions of relational membership indicators between the two nodes are independent. Under many social network settings, however, it is possible that certain known subgroups of people may have higher correlations in terms of their membership categories towards each other, and such prior information should be incorporated into the model. To this end, we introduce a new framework where individual Copula function is to be employed to model jointly the membership pairs of those nodes within the subgroup of interest. Under this framework, various Copula functions may be used to suit the scenario, while maintaining the membership's marginal distribution, as needed for modeling membership indicators with other nodes outside of the subgroup of interest. We will describe the model in detail and its sampling algorithm for both the finite and infinite (number of categories) case. Our experimental results shows a superior performance when comparing with the exisiting models on both the synthetic and real world datasets.

\end{abstract}

\section{Introduction}

Communities modeling is an emergent topic which has seen applications in various settings, including social-media recommendation \cite{tang2010community}, customer partitioning, discovering social networks, and partitioning protein-protein interaction networks \cite{girvan2002community, fortunato2010community}. Many models have been proposed in the last few years to address these problems; some earlier examples include \emph{Stochastic Blockmodel} \cite{doi:10.1198/016214501753208735} and \emph{Infinite Relational Model (IRM)} \cite{kemp2006learning}, both aiming to partition a network of nodes into different groups based on their pair-wise, directional binary observations.

The work of \emph{IRM} \cite{kemp2006learning} assumes that each node has one latent variable to directly indicate its community membership, dictated by a single distribution of communities. However, in many social network contexts, such representation may not well capture the complex interactions amongst the nodes, where multiple roles can possibly be played by a node. Two recent popular approaches to facilitate this phenomenon are: the \emph{latent feature model} and the \emph{latent class model}. The \emph{latent feature model} framework in \cite{hoff2002latent,hoff2005bilinear} assumes a latent real-valued feature vector for each node. The \emph{Latent Feature Relational Model (LFRM)} \cite{miller2009nonparametric} uses a binary vector to represent latent features of each node, and the number of features of all nodes can potentially be infinite using an Indian Buffet Process prior \cite{Griffiths05infinitelatent, griffiths2011indian}. The work in \cite{PalKnoGha12} further tries to uncover the substructure within each feature and uses the two nodes' ``co-active'' features while generating their interaction data.

Both models assume each node can associate with a set of latent features (i.e. communities). Their main difference is that, when deciding if or not a node has an interaction with another (i.e., relation), in the \emph{latent feature model}, all the associated communities from both nodes are considered. However, in the \emph{latent class model}, two communities are chosen, each to represent the one-way relationship between the two nodes.

A popular example of the \emph{latent class model} is \emph{MMSB} \cite{airoldi2008mixed}, where in order to facilitate the one-to-many relationships between a node and communities, each node has its own ``mixed-membership distribution'', in which its relationship with all other nodes is distributed from it. For nodes $i$ and $j$, after drawing their membership indicator pair, one then draws a final (directional) interaction from a so-called, ``role-compatibility matrix'' with row and column indexed by these pairs. A few variants are subsequently proposed from \emph{MMSB}, with examples including: \cite{koutsourelakis2008finding} which extends the \emph{MMSB} into the infinite communities case and  \cite{doi:10.1080/01621459.2012.682530} which uses the nested Chinese Restaurant Process \cite{Blei:2010:NCR:1667053.1667056} to build the communities' hierarchical structure and \cite{kim2012nonparametric} which incorporates the node's metadata information into \emph{MMSB}.

However, all of the above-mentioned \emph{MMSB}-type models make the assumption that for each pair of nodes, their membership indicator pairs were drawn independently, therefore limiting the way in which membership indicators can be distributed. Under many social network settings, however, certain known subgroups of people may have higher correlations towards each other in terms of their membership categories. For example, teenagers may have similar ``likes'' or ``dislikes'' on certain topics, compared with the views they may hold towards people of other age groups. Therefore, within a social networking context, we felt it important to incorporate such ``intra-subgroup correlation'' as prior information to the model. After introducing ``intra-subgroup correlations'', it is important that at the same time, we do not alter the distribution of membership indicators for the rest of the pairs of nodes.

Accordingly, in this paper, a Copula function \cite{nelsen2006introduction, mcneil2009multivariate} is introduced to \emph{MMSB}, forming a \emph{Copula Mixed-Membership Stochastic Blockmodel (cMMSB)}, for modeling the ``intra-subgroup correlations''. In this way, we can apply flexibly various Copula functions towards different subsets of pairs of nodes while maintaining the original marginal distribution of each of the membership indicators. We developed ways in which a bivariate Copula can be used for two distributions of indictors, enjoying infinitely possible values. Under the framework, we can incorporate different choices of Copula functions to suit the need of the application. With different Copula functions imposed on the different group of nodes, each of the Copula function's parameters will be updated in accordance with the data. What is more, we also give two analytical solutions to the calculation of the conditional marginal density to the two indicator variables, which plays a core role in our likelihood calculation and also gives new ideas on calculating a deterministic relationship between multiple variables in a graphical model.

In addition, there is also work on the time-varied relational model, for example, the stochastic blockmodel is used to capture the evolving community's behavior over time \cite{Yangml11}, which is addressed in \cite{conf/nips/IshiguroIUT10} by incorporating a time-varied Infinite Relational model; and the mixed-membership model is extended in \cite{xing2010state} with a dynamic setting. In this paper, we illustrate the \emph{cMMSB} by focusing on the static stochastic blockmodel; however, it can be equally extended into a time-varied setting.

The rest of the article is organized as follows: In Section \ref{sec_2}, we introduce the main model, including the notations and the details of our Copula-based \emph{MMSB}. We further provide two ``collapsed'' methods to do the probability calculation in Section \ref{sec_4}. In Section \ref{sec_5}, we show the experimental results based on our model, using both the synthetic and real-world social network data. In Section \ref{sec_6}, we conclude the paper by providing further discussion on our model.
%\vspace{-0.5\baselineskip}

\section{Copula Mixed Membership Stochastic Block model (\emph{cMMSB})} \label{sec_2}
\subsection{Notations \label{sec_21}}
We first present our notations and their meanings in Table \ref{table_22}. Readers who need a literature description can refer to Supplementary Material.
\begin{table}[htbp]
\caption{Notations for \emph{cMMSB}} \label{table_22}
\centering
\begin{tabular}{c|l}
\hline
$n$ & number of nodes\\
  \hline
$K$ & number of discovered communities\\
  \hline
$e_{ij}$ & directional, binary interactions\\
\hline
$\gamma, \alpha$ & concentration parameters for HDP\\
  \hline
$s_{ij}$ & sender's (from $i$ to $j$) membership indicator \\
  \hline
$r_{ij}$ & receiver's (from $j$ to $i$) membership indicator \\
  \hline
  \multirow{2}{*}{$\pi_i$} & mixed-membership distribution for node $i$,\\
& it generates $s_{\boldsymbol{i}1}, \cdots, s_{\boldsymbol{i}n}, r_{1\boldsymbol{i}}, \cdots, r_{n\boldsymbol{i}}$\\
  \hline
 $\pi_{ik}$ & the ``significance'' of community $k$ for  node $i$ \\
   \hline
 $B$ & role-compatibility matrix \\
  \hline
 $B_{k,l}$ & compatibilities between communities $k$ and $l$\\
  \hline
\multirow{2}{*}{$m_{k,l}$ } & number of links from community $k$ to $l$\\
 & i.e. $m_{ik}=\#\{ij: s_{ij}=k,r_{ij}=l.\}$\\
  \hline
\multirow{2}{*}{$m_{k,l}^{1}$}  & part of $m_{k,l}$ where the corresponding $e_{ij} = 1$\\
 & i.e. $m_{k,l}^1=\sum_{s_{ij}=k,r_{ij}=l} e_{ij}$\\
  \hline
\multirow{2}{*}{$m_{k,l}^{0}$}  & part of $m_{k,l}$ where the corresponding $e_{ij} = 0$\\
 & $m_{k,l}^0=m_{k,l}-m_{k,l}^1$\\
  \hline
\multirow{3}{*}{$N_{ik}$}  & number of times that a node $i$ has participated\\
& in community $k$ (either sending or receiving)\\
& i.e. $N_{ik}=\#\{j:s_{\boldsymbol{i}j}=k\}+\#\{j:r_{j\boldsymbol{i}}=k\}$\\
\hline
$\theta$ & parameter of Copula function\\
  \hline
  \end{tabular}
\end{table}

\subsection{Graphical model description} \label{sec_3}
\begin{figure}[htbp]
\centering
\includegraphics[scale=1, width = 0.4 \textwidth, bb = 145 512 345 670, clip]{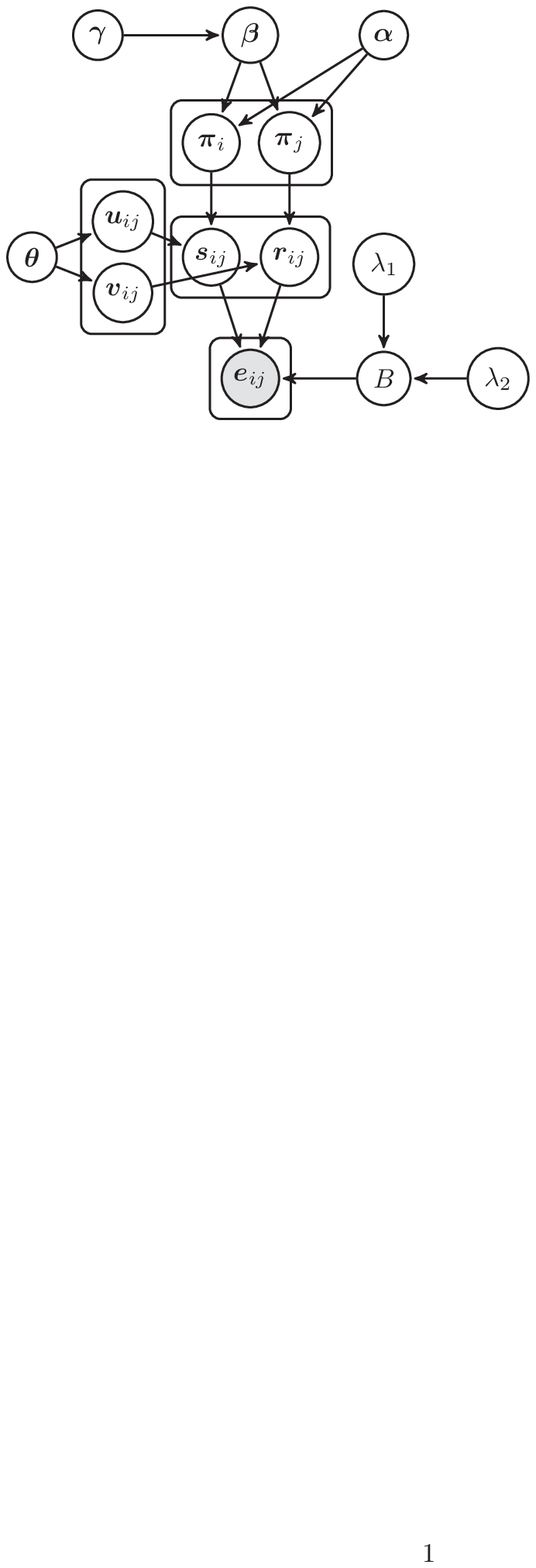}
\caption{Graphical model of \emph{Copula MMSB}} \label{fig_1}
\end{figure}
The corresponding generative model:
\begin{description}
\item{C$1$}: $\boldsymbol{\beta}\sim GEM (\gamma)$
\item{C$2$}: $\{\pi_i\}_{i=1}^n\sim DP (\alpha\cdot \boldsymbol{\beta})$
\item{C$3$}: $\left\{\begin{array}{ll}
(u_{ij}, v_{ij})\sim Copula(\theta), & g_{ij}=1;\\
u_{ij}, v_{ij}\sim U(0, 1), & g_{ij}=0. \\
\end{array}\right.$
\item{C$4$}: $s_{ij} = \Pi_i^{-1}(u_{ij}),r_{ij} = \Pi_j^{-1}(v_{ij})$
\item{C$5$}: $B_{k,l}\sim Beta (\lambda_1, \lambda_2), \forall k,l;$
\item{C$6$}: $\boldsymbol{e}_{ij}\sim Bernoulli (B_{s_{ij}, r_{ij}}).$
\end{description}
Here $g_{ij} = d$ in phase C$3$ denotes $s_{ij}, r_{ij}$ are in the $d^{\textrm{th}}$ correlated subgroup (in our case we mainly discuss $d=1$), while $g_{ij}=0$ means no subgroup correlation; in phase C$4$, $\Pi_i^{-1}(u_{ij}) = \{\min k: \sum_{q=1}^{k} \pi_{iq}\ge u_{ij}\}$, denoting for the interval of $\pi_i$ that $u_{ij}$ falls into, and the similar conclusion is to $\Pi_j^{-1}(v_{ij}) = \{\min k: \sum_{q=1}^{k} \pi_{jq}\ge v_{ij}\}$.

For the generative model shown in Figure \ref{fig_1}, phases C$1$-C$6$ describe the detail processes of our \emph{Copula mixed membership stochastic blockmodel}.

%as with the case of the classical \emph{MMSB} \cite{airoldi2008mixed}, the following conditions are used on the right hand side of the Figure.

%{\bf suggestion to delete} Our work is under the Hierarchical Dirichlet Process (HDP) \cite{teh2006hierarchical} setting, where each node $i$'s mixed-membership distribution $\{\pi_i\}$ is a weights vector on the importance of each communities and $\boldsymbol{\beta}$ is the common parent node for all $\{\pi_i\}_{i=1}^n$, as the graphical model in Figure \ref{fig_1}. A popular choice of prior for $\boldsymbol{\beta}$, when its number of components is fixed to an integer $k$ (i.e., there exist $k$ communities) is ``non-informative" symmetric Dirichlet distribution $(\boldsymbol{\beta}_1, \ldots, \boldsymbol{\beta}_k)\sim Dir(\gamma, \ldots, \gamma)$ \cite{airoldi2008mixed}.

\subsubsection{Mixed membership distribution modeling}
Phases C$1$-C$2$ are for the generation of each node's mixed membership distribution. As the number of communities $k$ is a vital issue in the mixed membership distribution, we consider two solutions here. The first is to use a fixed $k$: as the graphical model in Figure \ref{fig_1} shows, for all the mixed-membership distributions $\{\pi_i\}_{i=1}^n$, there is a common parent node $\boldsymbol{\beta}$, where $\boldsymbol{\beta}$ typically has a``non-informative" symmetric Dirichlet prior, i.e., $(\boldsymbol{\beta}_1, \ldots, \boldsymbol{\beta}_k)\sim Dir(\gamma, \ldots, \gamma)$ \cite{airoldi2008mixed}. The appropriate choice of $k$ is determined by model selection method, such as the BIC criterion \cite{schwarz1978estimating}, which is commonly used in \cite{airoldi2008mixed}\cite{xing2010state}.

The second solution is when the number of communities is uncertain, which is often the case in social network settings, where the usual approach is to use the Hierarchical Dirichlet Process (HDP) \cite{teh2006hierarchical} prior and $\boldsymbol{\beta}$ is to be distributed from a $\text{GEM} (\gamma)$, which is a stick-breaking construction of $\boldsymbol{\beta}$ \cite{sethuraman1991constructive} with $\boldsymbol{\beta}_k = u_k\prod_{l=1}^{k-1}(1-u_l), u_l\sim {Beta}(1, \gamma)$.

After obtaining their parent's node $\boldsymbol{\beta}$, we can sample our mixed-membership distribution $\{\pi_i\}$ independently from \cite{airoldi2008mixed,koutsourelakis2008finding}: $\pi_i\sim \left\{ \begin{array}{cc}
Dir(\alpha\cdot \boldsymbol{\beta}), & \textrm{fixed $k$}; \\
DP(\alpha\cdot \boldsymbol{\beta}), & \textrm{uncertain $k$}. \\
\end{array}
\right.$
For notational clarity, we concentrate our discussion on the uncertain $k$ case without explicitly mentioning its finite counterpart, as the fixed $k$ case can be trivially derived.

\subsubsection{Copula incorporated membership indicator pair}
Our main work of Copula incorporation into the membership indicator pair is displayed in phases C$3$-C$4$. A brief introduction to the Copula model is provided in Section 1 of the supplementary material. Readers unfamiliar with Copula functions are recommended to read this first.

Before entering the detail integration stage, we first the subgroup information our Copula functions' covering. Two cases are analysed here:
\begin{description}
\item{\bf Full correlation}: no subgroup information is given. We assume each pair of nodes of the whole dataset are using the same Copula function.  As we will see in the experiment section that, flexible modelling can still be achieved under this assumption, as parameters of a Copula can vary to support various form of relations.
\item{\bf Partial correlation}: the subgroup is pre-defined. This enables us to define a set of refined constraints in terms of intra-subgroup relations, when the context information is given.
\end{description}
For traditional \emph{MMSB}, the corresponding membership indicators within one pair $(s_{ij}, r_{ij})$ are independently sampled from their membership distributions, i.e., $s_{ij}\sim\pi_i, r_{ij}\sim\pi_j$. Using definition of $\{\Pi_i^{-1}(\cdot)\}_{i=1}^n$ from Section \ref{sec_3}, this is equivalently expressed as:
\begin{equation} \label{eq_5}
\begin{split}
& u_{ij}\sim U(0,1), v_{ij}\sim U(0,1);\\
& s_{ij} = \Pi_i^{-1}(u_{ij}),r_{ij} = \Pi_j^{-1}(v_{ij}).
\end{split}
\end{equation}
As discussed in the introduction, we are motivated by examples within social network settings, in which membership indicators from a node may well be correlated with other membership indicators in a subgroup of interest. People's interactions with each other within that subgroup may more likely (or less likely) belong to the same category, i.e., $(s_{ij}, r_{ij} )$ has higher (or lower) density in some regions of the discrete space $(1,2,\dots,\infty)^2$, which may not be well described by using only the two separate marginal distributions.

%Additionally, in some applications, one may possess some prior knowledge about the correlations of membership indicators for a` `subset'' of populations. As an example, we may know beforehand that within an organization, people from a certain subgroup may have probability of belonging (positive correlations) or not belonging (negative correlations) to the same category. At the same time, we do not possess any knowledge of correlations between this subgroup and others within the organization.

We propose a general framework by employing a Copula function to depict the correlation within the membership indicator pair. This is accomplished by the joint sampling of uniform variables $(u_{ij}, v_{ij})$ (in Eq. (\ref{eq_5}).) from the Copula function, instead of from two independent uniform distributions. More precisely, the membership indicator pair is obtained using:
\begin{equation} \label{eq_1}
\begin{aligned}[l]
\forall g_{ij}=1:(u_{ij},v_{ij})\sim Copula(u,v|\theta);\\
s_{ij} = \Pi_i^{-1}(u_{ij}),r_{ij} = \Pi_j^{-1}(v_{ij}).\\
\end{aligned}
\end{equation}
Using various Copula priors over the pair $(u_{ij}, v_{ij})$, we are able to express most appropriately the way in which the membership indicator pair $\{s_{ij}, r_{ij}\}$ is distributed, given the different scenarios we are facing. Taking the Gumbel Copula (with larger parameter values) \cite{nelsen2006introduction} as an instance, for certain membership indicator pairs ($g_{ij}=1$), it generates $(u_{ij},v_{ij})$ values more likely to have positive correlation, i.e.,  within $[0, 1]^2$ space, which promotes $s_{ij} = r_{ij}$. Also, the Gaussian Copula ($\theta=-1$) encourages the $(s_{ij},r_{ij})$ pair to be different.

\subsubsection{Binary observation modeling}
Phases C$5$-C$6$ model the binary observation, which directly follows the previous works \cite{doi:10.1198/016214501753208735}\cite{kemp2006learning} e.t.c.. Due to the beta-bernoulli conjugacy, $B$ can be marginalized out and the likelihood of binary observation is becomes as:
\begin{equation}
\Pr(\boldsymbol{e}|z, \lambda_1, \lambda_2)=\prod_{k,l}\frac{beta(m_{k,l}^1+\lambda_1, beta(m_{k,l}^0+\lambda_2)}{beta(\lambda_1, \lambda_2)}
\end{equation}
here $beta(\lambda_1,\lambda_2)$ denotes the beta function with parameters $\lambda_1$ and $\lambda_2$, $m_{k,l}^1$ and $m_{k,l}^0$ are defined in Table \ref{table_22}.

\section{Further discussion}
\subsection{Marginal conditional on $\pi$ or on $u,v$ only \label{sec_4}}
Let $K$ to be the discovered number of communities, a formal and concise representation of Eq. (\ref{eq_1}), i.e. the probability of $(s_{ij}, r_{ij})$, is:
\begin{equation} \label{eq_7}
\begin{split}
 \Pr&(s_{ij}, r_{ij})=\int_{\sum_{d=1}^{K+1}\pi_{jd}=1}\int_{\sum_{d=1}^{K+1}\pi_{id}=1}\int_{(u_{ij}, v_{ij})}\\
 &\cdot\boldsymbol{1}\left(s_{ij} = \Pi_i^{-1}(u_{ij}),r_{ij}= \Pi_j^{-1}(v_{ij})\right)\\
&\cdot dC(u_{ij}, v_{ij})dF(\pi_{i1}, \cdots,\pi_{iK+1})dF(\pi_{j1}, \cdots, \pi_{jK+1})\\
\end{split}
\end{equation}
Unfortunately, we cannot bring this \emph{total marginal density}, i.e., $\Pr(s_{ij}, r_{ij})$ to an analytical form without any integrals present. However, with some mathematical design, we found that with the explicit sampling of either $(u_{ij}, v_{ij})$ or $(\pi_{i}, \pi_{j})$, it is possible to obtain a marginalised conditional density in which $s_{ij}, r_{ij}$ is conditioned on either $(u_{ij}, v_{ij})$ or $(\pi_{i}, \pi_{j})$, but not both. Additionally, having a set of variables ``collapsed'' from sampling results in a faster mixing on Markov chains \cite{liu1994collapsed}.

\subsubsection{Marginal conditional on $\pi$ only ($\textrm{\emph{cMMSB}}^{\pi}$)}
Here, we define $p_{ij}^{kl}(\pi_i, \pi_j) \equiv \Pr(s_{ij}=k, r_{ij}=l|\pi_i, \pi_j, \theta_d), \forall g_{ij}=1$, and let $C( u_{ij},v_{ij}|\theta_d)$  be the chosen Copula cumulative distribution function (c.d.f.) with parameter $\theta_d$. Given the explicit values of $\pi_i, \pi_j$, we can integrate over all $u_{ij},v_{ij}$ to compute the probability mass of the indicator pair $(s_{ij}=k, r_{ij}=l), k,l\in\{1, \cdots,K+1\}$:
\begin{equation} \label{eq_2}
\begin{split}
& p_{ij}^{kl}(\pi_i, \pi_j)=\int_{\hat{\pi}_{i}^{k-1}}^{\hat{\pi}_{i}^{k}}\int_{\hat{\pi}_{j}^{l-1}}^{\hat{\pi}_{j}^{l}} d C(u,v|\theta_d) \\
= &C(\hat{\pi}_{i}^{k}, \hat{\pi}_{j}^{l})+C(\hat{\pi}_{i}^{k-1}, \hat{\pi}_{j}^{l-1})-C(\hat{\pi}_{i}^{k}, \hat{\pi}_{j}^{l-1})-C(\hat{\pi}_{i}^{k-1}, \hat{\pi}_{j}^{l})
\end{split}
\end{equation}
Here $\hat{\pi}_i^k=\left\{\begin{array}{cc}
0, & k=0;\\
\sum_{q=1}^{k} \pi_{iq}, & k>0\\
\end{array}\right.$. Since $\{\pi_i\}_{i=1}^n$ are piecewise functions, we can easily calculate this ``rectangular'' area. In other cases of $\{g_{ij}=0\}$, i.e., interaction data $e_{ij}$ outside the correlated subgroup, we have $p_{ij}^{kl}(\pi_i, \pi_j)=\pi_{ik}\pi_{jl}$.

It is noted that, using the properties of a Copula function, the marginal distributions of $\Pr(s_{ij}=k, r_{ij}=l|\pi_i, \pi_j, \theta_d)$ remain $\pi_i$ and $\pi_j$ respectively, which becomes that of:
\begin{equation}
\begin{aligned}[l]
\sum_{l=1}^{K+1} \Pr(s_{ij}=k, r_{ij}=l|\pi_i, \pi_j, \theta_d)=\pi_{ik};\\
\sum_{k=1}^{K+1} \Pr(s_{ij}=k, r_{ij}=l|\pi_i, \pi_j, \theta_d)=\pi_{jl}.\\
\end{aligned}
\end{equation}
%
%We shall notice here that each membership indicator pair $\{(u_{ij}, v_{ij})\}_{i,j}$ are pairwise independent as their common parent nodes, i.e. $\{\pi_i\}_{i=1}^n$ and $C(u,v|\theta_d)$, are substantialized.
%
\subsubsection{Marginal conditional on $u,v$ only ($\textrm{\emph{cMMSB}}^{uv}$)}
An alternative ``collapsed'' sampling method is to integrate over $\{\pi_i\}_{i=1}^n$ while we explicitly sample the values of $\{(u_{ij}, v_{ij})\}_{i,j}$.

From Eq. (\ref{eq_7}), given $\{(u_{ij}, v_{ij})\}_{i,j}$'s values, the probabilities $s_{ij}= k$ and $r_{ij}=l$ can be computed independently. The Copula function leaves marginal distributions of $s_{ij}$ and $r_{ij}$ invariant, which remains the same as the classical \emph{MMSB}, i.e., $\pi_i |\alpha, \beta, \{N_{ik}^{-ij}\}_{k=1}^K\sim Dir(\alpha\beta_1+N_{i1}^{-ij}, \cdots, \alpha\beta_K+N_{iK}^{-ij},\alpha\beta_{K+1})$. Therefore, having the knowledge of $F(\pi_i |\alpha, \beta, \{N_{ik}^{-ij}\}_{k=1}^K)$, given $u_{ij}$, our calculation of $\Pr(s_{ij}=k)$ is equal to computing the probability of $u_{ij}$ falling in $\pi_i$'s $k^{\textrm{th}}$ interval, i.e. $\Pr(\sum_{d=1}^{k-1} \pi_{id}\le u_{ij}<\sum_{d=1}^{k} \pi_{id})$ (similar case with $v_{ij}$ to $\pi_{jl}$). This can be obtained from the fact that the set $\{u_{ij}\in[0,1]|\sum_{d=1}^{k-1} \pi_{id}\le u_{ij}\}$ can be decomposed into two {\bf \emph{disjoint}} sets:
{
\begin{equation}
\begin{split}
& \{u_{ij}\in[0,1]|\sum_{d=1}^{k-1} \pi_{id}\le u_{ij}\}\\
 = &\{u_{ij}\in[0,1]|\sum_{d=1}^{k-1} \pi_{id}\le u_{ij}<\sum_{d=1}^{k} \pi_{id}\}\\
 &\cup\{u_{ij}\in[0,1]|\sum_{d=1}^{k} \pi_{id}\le u_{ij}\}
\end{split}
\end{equation}
}
where $\sum_{d=1}^k \pi_{id}\sim Beta(\sum_{d=1}^k \alpha\beta_d+N_{id}, \sum_{d=k+1}^{K+1} \alpha\beta_d+N_{id})$. (A similar result was also found in page 10 of \cite{teh2006hierarchical}). Therefore, we get:

\begin{equation}
\begin{split}
& \Pr(\sum_{d=1}^{k-1} \pi_{id}\le u_{ij}<\sum_{d=1}^{k} \pi_{id})\\
=&\Pr(\sum_{d=1}^{k-1} \pi_{id}\le u_{ij})-\Pr(\sum_{d=1}^{k} \pi_{id}\le u_{ij})\\
=& I_{u_{ij}}(h_{i}^{k-1}, \hat{h}_{i}^{k-1})-I_{u_{ij}}(h_{i}^{k}, \hat{h}_{i}^{k})
\end{split}
\end{equation}

Here $h_{i}^{k} = \sum_{d=1}^k \alpha\beta_d+N_{id}, \hat{h}_{i}^{k} = \sum_{d=k+1}^{K+1} \alpha\beta_d+N_{id}$; $I_u(a, b)$ denotes the Beta c.d.f. value with parameter $a, b$ on $u$. The existence and non-negativity of $I_{u_{ij}}(u_{k-1}, \hat{u}_{k-1})-I_{u_{ij}}(u_k, \hat{u}_k)$ is guaranteed by the fact that $\{u_{ij}\in[0,1]|\sum_{d=1}^{k} \pi_{id}\le u_{ij}\}\subseteq\{u_{ij}\in[0,1]|\sum_{d=1}^{k-1} \pi_{id}\le u_{ij}\}$ on the same $\pi_{i}$.

\subsection{Relations with classical \emph{MMSB}}

A bivariate independence Copula function, i.e. $C(u,v)=uv$, is a uniform distribution on the region of $[0,1]\times[0,1]$. Under the case of ``marginal conditional on $\pi$ only'', Eq. (\ref{eq_2}) then becomes that of $p_{ij}^{kl}(\pi_i, \pi_j) = \Pr(s_{ij}=k, r_{ij}=l|\pi_i, \pi_j)=\int_{\hat{\pi}_{i}^{k-1}}^{\hat{\pi}_{i}^{k}}\int_{\hat{\pi}_{j}^{l-1}}^{\hat{\pi}_{j}^{l}}\cdot 1 dudv= \pi_{ik}\cdot\pi_{jl}$. Under the case of ``marginal conditional on $u,v$ only'', as $\{u_{ij}, v_{ij}\}$ are independently uniform distributed, the equation $\int_{u_{ij}}\Pr(\sum_{d=1}^{k-1} \pi_{id}<u_{ij}\le\sum_{d=1}^{k} \pi_{id})du_{ij}=\sum_{d=1}^{k} \pi_{id}-\sum_{d=1}^{k-1} \pi_{id}=\pi_{ik}$. (A similar result also holds for $v_{ij}$.) All these results are identical to that of the classical \emph{MMSB}. In a sense, our model can be viewed as a generalization of \emph{MMSB}.

In addition, for most Copula functions, a certain choice of parameters will result in the function equaling or approaching that of the independence Copula. As an example, when Gumbel \cite{nelsen2006introduction} Copula is used, which has its c.d.f. defined as:
\begin{equation}
C(u,v)=\exp\left[-\left((-\ln u)^{\theta}+(-\ln v)^{\theta}\right)^{\frac{1}{\theta}}\right]
\end{equation}
where $\theta\in[1,\infty)$. For $\theta = 1$, it becomes that of the independence Copula. Our experiments show that when the data are generated using independence Copula (i.e., classical \emph{MMSB}), the recovered Gumbel Copula's parameter has a high probability of around $1$.

\subsection{Relations with Dependent Dirichlet Process}
As we should note that, since the proposal of \emph{dependent dirichlet process (DDP)} \cite{maceachern1999dependent}, a variety of \emph{DDP} models were developed, including a recent  Poission Process perspective \cite{lin2010construction} and its variants \cite{lin2012coupling}\cite{foti2013unifying}\cite{CheRaoBun2013a}.

From the dependency modeling perspective, our Copula incorporated work achieves a similar goal that of \emph{DDP}. However, the \emph{DDP}-type works concentrate on the intrinsic relations between multiple Dirichlet Processes. In our work, however, we assume Dirichlet Processes themselves are independent. The dependency is introduced at the (Discrete) realizations of the multiple DPs, which are the membership indicators. Therefore, making it feasible to use Copula to model the dependency between each pair of membership indicators. This obviously can not be achieved at the DP level, as one's relations with every other nodes share the same DP.

\subsection{Computational complexity analysis}
We estimate the computational complexity for each graphical model and present the result in Table \ref{table_0}. Comparing to the classical models (especially the \emph{MMSB}), our $\textrm{\emph{cMMSB}}^{\pi}$ involves an additional $\mathcal{O}(Kn)$ term which refers to the sampling of the mixed membership distributions. Note that the computation time varies for different Copulas. $\textrm{\emph{cMMSB}}^{uv}$ requires an extra $\mathcal{O}(n^2)$ term  for the $u,v$'s sampling for each membership indicator. Each operation requires a beta c.d.f. in a tractable form.\begin{table}[htbp]
\caption{Computational Complexity for different Models} \label{table_0}
\centering
\begin{tabular}{c|c}
\hline
Models & Computational complexity \\
  \hline
\emph{IRM}  &  $\mathcal{O}(K^2n)$ \cite{PalKnoGha12}\\
\emph{LFRM}  &  $\mathcal{O}(K^2n^2)$ \cite{PalKnoGha12}\\
\emph{MMSB}  &  $\mathcal{O}(Kn^2)$ \cite{kim2012nonparametric}\\
  $\textrm{\emph{cMMSB}}^{\pi}$  &  $\mathcal{O}(Kn^2+Kn)$ \\
  $\textrm{\emph{cMMSB}}^{uv}$  &  $\mathcal{O}(Kn^2+n^2)$ \\
  \hline
  \end{tabular}
\end{table}

\begin{table*}[htbp]
\caption{Different models' performance (Mean $\mp $ Standard Deviation) on Synthetic data in Full Correlation case. An interesting part of our results is that we find the \emph{IRM} is slightly better than the \emph{LFRM} and \emph{MMSB}, we explain this to the ``blockness'' of the synthetic data. In terms of \emph{train error}, our model is comparative to other \emph{MMSB}-type models, which in general outperforms \emph{IRM} and \emph{LFRM}. On the predictivity measures on \emph{test error}, \emph{test log likelihood} and \emph{AUC}, both our $\textrm{\emph{cMMSB}}^{\pi}$ and $\textrm{\emph{cMMSB}}^{uv}$ outperform all other \emph{MMSB} and non-\emph{MMSB} benchmarks.} \label{table_1}
\centering
\begin{threeparttable}
\begin{tabular}{c|cccc}
  \hline
 & {\small \emph{Train error}} & {\small \emph{Test error}} & {\small\emph{ Test log likelihood}} & {\small\emph{ AUC}} \\
  \hline
  \emph{IRM} & $    0.1044 \mp     0.0099$ & $    0.1047 \mp     0.0126$ & $  -90.0597 \mp     8.2165$ & $    0.8735 \mp     0.0256$    \\
  \emph{LFRM} &$    0.0944 \mp     0.0029$ & $    0.1132 \mp     0.0141$ & $  -99.9926 \mp    10.8915$& $    0.8724 \mp     0.0230$  \\
  \emph{MMSB}  &$  \boldsymbol{  0.0236 \mp     0.0002}$   & $    0.1248 \mp     0.0004$    & $ -104.107 \mp     0.265$   & $    0.8510 \mp     0.0011$  \\
  \emph{iMMM} &$    0.0266 \mp     0.0002$  &  $    0.1208 \mp     0.0003$    &$ -101.497 \mp     0.203$  & $    0.8619 \mp     0.0008$ \\
  $\textrm{\emph{cMMSB}}^{\pi}$ &$    0.0332 \mp     0.0003$  &  $  \boldsymbol{  0.0884 \mp     0.0002}$    &$ \boldsymbol{ -82.625 \mp     0.128}$  & $ \boldsymbol{   0.8897 \mp     0.0003}$   \\
  $\textrm{\emph{cMMSB}}^{uv}$ &$    0.0423 \mp     0.0002$  &  $0.0932 \mp     0.0001$   &$ -85.951 \mp     0.084$ & $    0.8889 \mp     0.0010$   \\
    $\textrm{\emph{cMMSB}}^{\pi}$(P)\tnote{1} &$    0.0342\mp     0.0005$  &  $   0.0891 \mp     0.0004$    &$ -83.264 \mp     0.105$  & $   0.8940 \mp     0.0012$   \\
  $\textrm{\emph{cMMSB}}^{uv}$(P)\tnote{1} &$    0.0497 \mp     0.0010$  &  $0.0908 \mp     0.0013$   &$ -83.124 \mp     0.046$ & $    0.8946 \mp     0.0073$   \\
  \hline
  \end{tabular}
        \begin{tablenotes}
        \footnotesize
        \item[1] This is under the situation of Partial Correaltion, i.e., we are using two Copula functions in different subgroups.
      \end{tablenotes}
\end{threeparttable}
\end{table*}

\section{Experiments} \label{sec_5}
%\vspace{-1\baselineskip}
Here, our \emph{Copula-MMSB}'s performance is compared with the classical \emph{mixed-membership stochastic blockmodel (MMSB)}-type methods, including the original \emph{MMSB} \cite{airoldi2008mixed} and the \emph{infinite mixed-membership model (iMMM)} \cite{koutsourelakis2008finding}. Additionally, we also compare it with other non-MMSB approaches including the \emph{infinite relational model (IRM)} \cite{kemp2006learning} and the \emph{latent feature relational model (LFRM)} \cite{miller2009nonparametric}.

We have independently implemented the above benchmark algorithms to the best of our understanding. In order to provide common ground for all comparisons, we made the following small variations to these algorithms: (1) In \emph{iMMM}, instead of having an individual $\alpha_i$ value for each $\pi_i$ as used in the original work, we used a common $\alpha$ value for all the mixed-membership distributions $\{\pi_i\}_{i=1}^n$; (2) In \emph{LFRM} \cite{miller2009nonparametric}'s implementation, we do not incorporate the metadata information into the interaction data's generation, but to use only the binary interaction information.

\subsection{Synthetic data}
We first perform the synthetic data exploration as a pilot study. In addition to the ones associated with the Copula function, the rest of the variables are generated in accordance with \cite{airoldi2008mixed, newman2004finding}. We used $n = 50$, and hence $E$ is a $50\times 50$ asymmetric, binary matrix. The parameters are setup such that $50$ nodes were partitioned into $4$ subgroups, with each subgroup having $20, 13, 9,8$ number of nodes, respectively. The mixed-membership distribution of each group and the whole role-compatibility matrix are displayed in Figure \ref{fig_2} and Figure \ref{fig_3}, respectively. Thus, the generated synthetic data forms as one {\bf block diagonal matrix}, with the outliers existed.

\begin{figure} [htbp]
\centering
    \begin{tabular}{|cccc|}
    0.9 & 0.1 & 0 & 0 \\
    0 & 0.9 & 0.1 & 0 \\
    0.1 & 0.05 & 0.85 & 0 \\
    0.1 & 0.05 & 0.05 & 0.8 \\
    \end{tabular}
    \caption{Mixed-membership distribution} \label{fig_2}
\end{figure}
\begin{figure}[htbp]
    \centering
    \begin{tabular}{|cccc|}
    0.95 & 0.05 & 0 & 0 \\
    0.05 & 0.95 & 0.05 & 0\\
    0.05 & 0 & 0.95 & 0\\
    0 & 0.05 & 0 & 0.95\\
    \end{tabular}
    \caption{Role-compatibility matrix} \label{fig_3}
\end{figure}

\subsubsection{Full correlation - Single Copula on all nodes in link prediction} \label{sec_7}
We incorporated a single Gumbel Copula (with parameter $\theta$ = 3.5) on every interaction to generate all membership indicator pairs. This model is tested against link prediction performance. We use a ten-folds cross-validation on the interaction data and show the corresponding comparisons in Table \ref{table_1}. We provide definitions for \emph{train error}, \emph{test error}, \emph{test log likelihood} and \emph{AUC} in Section 4 of the supplementary material.

\begin{figure*}[htbp]
\centering
\includegraphics[scale=1, width = 0.6182 \textwidth, bb = 99 335 492 553, clip]{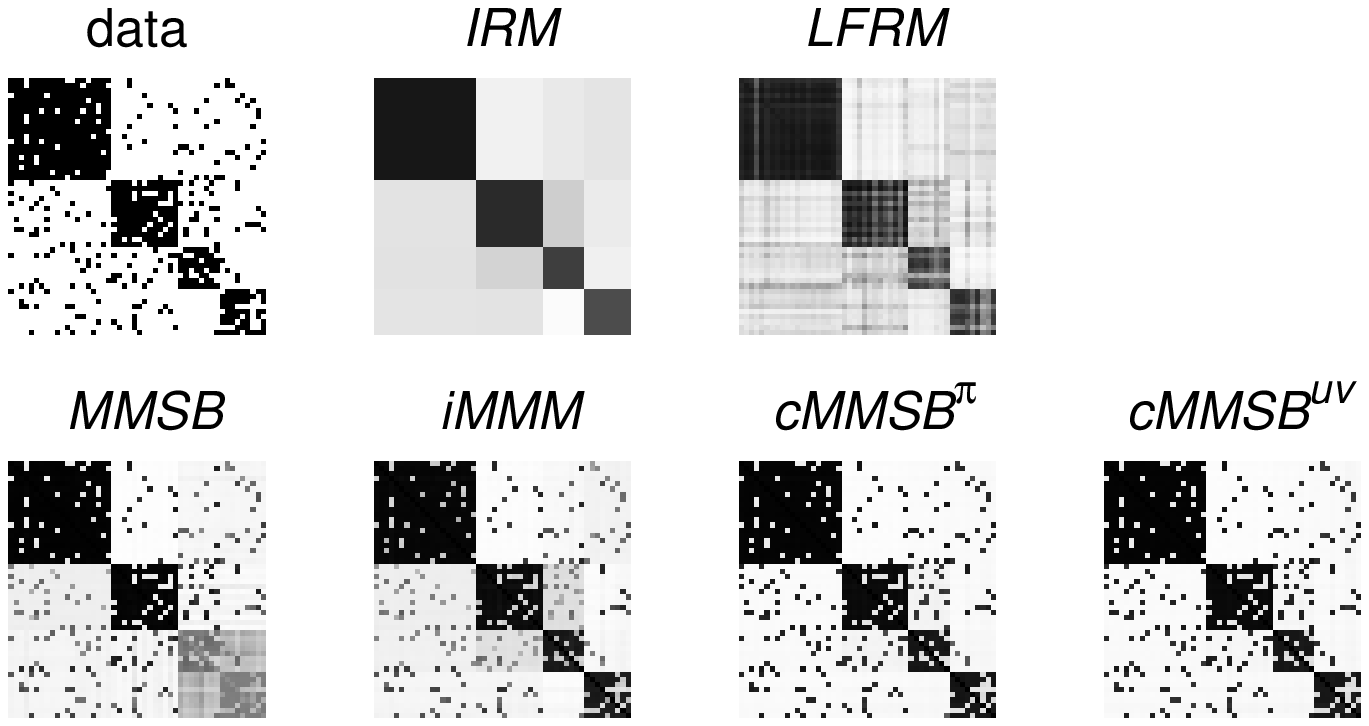}
\caption{Comparison of the models' posterior predictive distribution on the training data. The original data is a block diagonal matrix, with some outliers existed as the black points. For the \emph{IRM} model, its result is composed of rectangular zones. one value is presented in each rectangular. This simplified and ``smoothed'' version is due to the single membership representation for one node, it can not distinguish the random distributed points. Comparing to this, the \emph{LFRM} provides a larger amount of values to select from. This enable the model to place different values on one rectangular zone, especially each node is meant to be line-shaped colors, which is in consistent with the one latent features vector for one node representation. However, it still fails to detect the random points. Then comes the \emph{MMSB} and \emph{iMMM} in that they have successfully captured the random points. What is more, we find our $\textrm{\emph{cMMSB}}^{\pi}$ and $\textrm{\emph{cMMSB}}^{uv}$ models partition the relational data the best. } \label{fig_4}
\end{figure*}

Another interesting comparison is the posterior predictive distribution on the train data, as the results shown in Figure \ref{fig_4}. The corresponding posterior predictive distribution is calculated as the average value of the effective samples, the second half of the samples in one chains as we set the first half being the ``burn in'' stage. The darker of the pointer stands for the larger value close to 1, and vice versa. %For the \emph{IRM} model, its result is composed of rectangular zones. one value is presented in each rectangular. This simplified and ``smoothed'' version is due to the single membership representation for one node, it can not distinguish the random distributed points. Comparing to this, the \emph{LFRM} provides a larger amount of values to select from. This enable the model to place different values on one rectangular zone, however, it still fails to detect the random points. Then comes the \emph{MMSB} and \emph{iMMM} in that they have successfully captured the random points.
%Additionally, we also ran 10 Markov Chain boxplot of Frobenius norm, i.e. ``entrywise'' $L_2$ norm between the discovered role-compatibility matrix and the ground-truth in the left part of Figure 4, which shows a better performance than the classical \emph{MMSB} and \emph{iMMM}.
\subsubsection{Partial correlation - Multiple Copulas in subgroup structure}
We also have an additional test case and integrate two Gumbel Copula functions in the modelling. The first 20 nodes are forming a correlated subgroup and share one Copula function, while the other Copula function is applied on the rest of the interactions. The performance on the Full correlation data is shown in Table \ref{table_1}.

While using this model on a Partial correlation dataset, we get the $95\%$ Confidence Interval for both of the recovered $\theta_1$ and $\theta_2$ displayed in Table \ref{table_2}. We can see that our model can distinguish between the correlated and independent case, where the recovered value of $\theta_2$ is much closer to 1.

\begin{table}[htbp]
\caption{$\theta$'s $95\%$ Confidence Interval} \label{table_2}
\centering \small
\begin{tabular}{c|ccc}
\hline
Models & s-cMMSB & s-ciMMM & Ground-truth \\
  \hline
$\theta_1$  & $4.19 \mp 0.91$ &$3.23 \mp 1.22$ & 3.5 \\
  $\theta_2$ & $1.42 \mp 0.23$& $2.39 \mp 0.48$ & 1.0 \\
  \hline
  \end{tabular}
\end{table}

%\subsubsection{Computation cost}
%
%Since we have incorporated the Copula function, its running time is tested against classical \emph{MMSB}. When assuming a single Copula, we find our method is around 1.5 times more computational than the classical algorithms. A detailed comparison is shown in Section 3 of the supplementary material.
%
%\subsubsection{Hyper-parameters' initialization discussion}

\subsection{Real world datasets' link prediction}
%We selected five real-life datasets for benchmark testing and give their detailed information, including the data source, number of nodes, link number and link types in Section 5 of the supplementary materials. Table \ref{table_real} shows the \emph{AUC} performance of link prediction on these four datasets, with the same setting as Section \ref{sec_7}. From these reported statistics, we can see that our methods ($\textrm{\emph{cMMSB}}^{\pi}$, $\textrm{\emph{cMMSB}}^{uv}$) obtain the best performance in the last 3 datasets, of all other models. Even in the reality dataset, our models' performance is quite comparable as it is close to the best one.
We analyse three real world datasets here: the NIPS Co-authorship dataset, the MIT Reality Mining dataset and the lazega-lawfirm dataset. As the predict ability is one important property of the model, we use a ten-folds cross-validation to complete this task, where we randomly select one out of ten for each node's link data as test data and the others as training data. The criteria in evaluating this predict ability includes the train error ($0-1$ loss), the test error ($0-1$ loss), the test log likelihood and the AUC (Area Under the roc Curve) score, where detail derivations of these values can refer to the Supplementary Material. Table \ref{table_3} shows the detail results.

%\begin{table*}[htbp]
%\caption{\emph{AUC} performance (Mean $\mp $ Standard Deviation)} \label{table_real}
%\centering
%\begin{tabular}{c|ccccc}
%  \hline
%& {\small NIPS co-authorship} & {\small MIT reality} & {\small Lazega-law} & {\small Food Web } & {\small Wiki-talk } \\
%  \hline
%  \emph{IRM} & $    0.8846 \mp     0.0009$  & $    0.8047 \mp     0.0003$ & $  0.7794 \mp  0.0002$   &$    0.7401 \mp     0.0012$   & $    0.7802 \mp     0.0010$      \\
%  \emph{LFRM} & $    0.8858 \mp     0.0427$  &$    0.8503 \mp     0.0094$  & $    0.7987 \mp     0.0123$   & $    0.7590 \mp     0.0292$   & $    0.7891 \mp     0.0142$    \\
%  \emph{MMSB}  & $    0.8869 \mp     0.0295$ &$    0.8839 \mp     0.0064$  & $    0.7539 \mp     0.0105$     & $    0.7408 \mp     0.0290$    & $    0.7940 \mp     0.0113$    \\
%  \emph{iMMM}  & $ \boldsymbol{   0.9341 \mp     0.0390}$  &$   \boldsymbol{  0.8994 \mp     0.0061}$    &  $    0.7974 \mp     0.0118$    &$    0.7692 \mp     0.0171$  & $    0.8220 \mp     0.0177$  \\
%  $\textrm{\emph{cMMSB}}^{\pi}$ & $    0.9204 \mp     0.0139$  &$    0.8879 \mp     0.0423$  &  $  \boldsymbol{  0.8084 \mp     0.0622}$    &$ \boldsymbol{ 0.7872 \mp     0.0146}$  & $ 0.8140 \mp     0.0332$  \\
%  $\textrm{\emph{cMMSB}}^{uv}$ & $    0.9137 \mp     0.0314$  &$    0.8843 \mp     0.0213$  &  $0.7932 \mp     0.0390$   &$  0.7772 \mp     0.0544$ & $  \boldsymbol{  0.8303 \mp     0.0982}$   \\
%  \hline
%  \end{tabular}
%\end{table*}

\begin{table*}[htbp]
\caption{Different models' performance (Mean $\mp $ Standard Deviation) on Real world datasets. From these reported statistics, we can see that our methods ($\textrm{\emph{cMMSB}}^{\pi}$, $\textrm{\emph{cMMSB}}^{uv}$) obtain the best performance in these 3 datasets, amongest all other models.}
 \label{table_3}
\centering
\begin{tabular}{c|c|cccc}
  \hline
dataset & & {\small \emph{Train error}} & {\small \emph{Test error}} & {\small\emph{ Test log likelihood}} & {\small\emph{ AUC}} \\
  \hline
&  \emph{IRM} & $    0.0317 \mp     0.0004$ & $    0.0423 \mp     0.0014$ & $ -135.0467 \mp     7.3816$ & $    0.8901 \mp     0.0162$ \\
&  \emph{LFRM} &$    0.0482 \mp     0.0794$ & $    0.0239 \mp     0.0735$ & $ -105.2166 \mp   179.5505$ & $    0.9348 \mp     0.1667$ \\
NIPS & \emph{MMSB}  &$    0.0132 \mp     0.0042$ & $    0.0301 \mp     0.0064$ & $  -86.2134 \mp     10.1258$ & $    0.9524 \mp     0.0215$ \\
co-author & \emph{iMMM} &$  \boldsymbol{0.0061 \mp 0.0019}$ & $    0.0253 \mp     0.0035$ & $  -83.4264 \mp     9.4293$ & $    0.9574 \mp     0.0155$ \\
 & $\textrm{\emph{cMMSB}}^{\pi}$ &$    0.0066 \mp     0.0038$&$  \boldsymbol{   0.0231 \mp     0.0043}$&$  -83.4261 \mp     9.4280$&$    0.9569 \mp     0.0159$\\
 & $\textrm{\emph{cMMSB}}^{uv}$ &$    0.0097 \mp     0.0047$&$    0.0240 \mp     0.0065$&$  \boldsymbol{ -83.4257 \mp     9.4292}$&$  \boldsymbol{ 0.9581 \mp  0.0153}$ \\
  \hline
&  {\emph{IRM}} &  $    0.0627 \mp     0.0002$ & $    0.0665 \mp     0.0004$ & $ -133.8037 \mp     1.1269$ & $    0.8261 \mp     0.0047$ \\
&  {\emph{LFRM}} & $    0.0397 \mp     0.0017$ & $    0.0629 \mp     0.0037$ & $ -143.6067 \mp    10.0592$ & $    0.8529 \mp     0.0179$ \\
MIT & {\emph{MMSB}}  & $    0.0263 \mp     0.0105$  & $    0.0716 \mp     0.0043$  &  $ -129.4354 \mp     7.6549$   & $    0.8561 \mp     0.0176$  \\
realtiy & {\emph{iMMM}} & $    0.0297 \mp     0.0055$  & $    0.0625 \mp     0.0015$  &  $ -126.7876 \mp     3.4774$   & $    0.8617 \mp     0.0124$  \\
 & $\textrm{\emph{cMMSB}}^{\pi}$ &$   \boldsymbol{  0.0246 \mp     0.0016}$  & $    0.0489 \mp     0.0016$  &  $ -125.3876 \mp     3.2689$   & $   \boldsymbol{  0.8794 \mp     0.0159}$  \\
 & $\textrm{\emph{cMMSB}}^{uv}$ &$    0.0283 \mp     0.0035$  & $  \boldsymbol{   0.0438 \mp     0.0015}$  &  $  \boldsymbol{-123.3876 \mp     3.1254}$   & $    0.8738 \mp     0.0364$  \\
  \hline
& {\emph{IRM}} & $    0.0987 \mp     0.0003$ & $    0.1046 \mp     0.0012$ & $ -201.7912 \mp     3.3500$
& $    0.7056 \mp     0.0167$ \\
&  {\emph{LFRM}} & $    0.0566 \mp     0.0024$ & $    0.1051 \mp     0.0064$ & $ -222.5924 \mp    16.1985$ & $    0.8170 \mp     0.0197$ \\
Lazega & {\emph{MMSB}}  & $    0.0391 \mp     0.0071$   & $    0.0913 \mp     0.0030$  &  $ -212.1256 \mp     3.2145$   & $    0.7989 \mp     0.0102$  \\
lawfirm & {\emph{iMMM}} & $    0.0487 \mp     0.0068$   & $    0.1096 \mp     0.0026$  &  $ -202.7148 \mp     5.3076$   & $    0.8074 \mp     0.0141$  \\
 & $\textrm{\emph{cMMSB}}^{\pi}$ &$   \boldsymbol{    0.0246 \mp     0.0050}$   & $   \boldsymbol{  0.1023 \mp     0.0056}$  &  $   \boldsymbol{ -201.0154 \mp     5.2167}$   & $  \boldsymbol{ 0.8273 \mp     0.0148}$  \\
 & $\textrm{\emph{cMMSB}}^{uv}$ &$    0.0276 \mp     0.0043$   & $    0.1143 \mp     0.0019$  &  $ -204.0289 \mp     9.5460$   & $    0.8215 \mp     0.0167$  \\
  \hline
  \end{tabular}
\end{table*}

\subsubsection{NIPS co-authorship dataset}
We use the co-authorship as a relation from the proceeding of the \emph{Neural Information Processing Systems} (NIPS) conference for the years 2000-2012. As the sparsity of the co-authorship, we observe the authors' activities in all the 13 years (i.e. regardless of the time factor) and set the relational data being 1 if the two corresponding authors have co-authored for no less than 2 papers, which is to remove the co-authors' randomness. Further, the author with less than 4 relationships with others are manually eliminated. Thus, a $92\times92$ symmetric, binary matrix is obtained.

On this dataset, no pre-defined subgroup information is obtained in advance. Thus, we consider it as full-correlation case and use one Gumbel Copula function in modelling all the interactions.

\subsubsection{MIT Reality dataset}
From the MIT Reality Mining \cite{Eagle:2006:RMS:1122739.1122745}, we have used the subjects' proximity dataset, where weighted links are indicating the average proximity from one subject to another at work. We then``binarize'' the data, in which we set the proximity value larger than 10 minutes per day as 1, and 0 otherwise. Therefore, a $94\times 94$ asymmetric, binary matrix is obtained.

The dataset are roughly divided into four groups: Sloan Business School students (Sloan), Lab faculty, senior students with more than 1 year in the lab and junior students. In our experiment, we have only applied the Gumbel Copula function to the Sloan portion of the students to encourage similar mixture membership indicators.

\subsubsection{Lazega Law dataset}
The lazega-lawfirm dataset \citep{lazega2001collegial} is obtained from a social network study of corporate located in the northeastern part of U.S. in 1988 - 1991. The dataset contains three different types of relations: co-work network, basic advice network and friendship network, among the 71 attorneys, of which the element are labeled as $1$ (exist) or $0$ (absent).

Since no subgroup information is obtained in this dataset, we use one Gumbel Copula function on the whole group and show the corresponding result in Table \ref{table_3}.

\section{Conclusions} \label{sec_6}
%\vspace{-1\baselineskip}
In this paper, we have proposed a new framework to realistically describe the intra-subgroup correlations in a \emph{Mixed-Membership Stochastic Blockmodel}. The key to the model is the introduction of the Copula function, which represents the correlation between the pair of membership indicators, while keeping the membership indicators' marginal distribution invariant. The results show that, using both synthetic and real data, our Copula-incorporated \emph{MMSB} is effective in learning the community structure and predicting the missing links, when information within a subgroup is known.

In terms of inference, our main contribution is to have obtained an analytical solution to both of the conditional marginal likelihoods to the two indicator variables $(s_{ij}, r_{ji})$, given \emph{either} the indicator distributions $\pi_i, \pi_j$ \emph{or} the bivariate Copula variables $u_{ij},v_{ij}$.

% Can use something like this to put references on a page
% by themselves when using endfloat and the captionsoff option.
%\ifCLASSOPTIONcaptionsoff
%  \newpage
%\fi

{\small
\bibliography{example_paper}
\bibliographystyle{icml2014}
}
\end{document}